\documentclass[11pt,a4paper]{article}
\usepackage[hyperref]{acl2017}
\usepackage{times}
\usepackage{latexsym}
\usepackage{graphicx}
\usepackage{amsmath}

\usepackage{url}

\aclfinalcopy 

\title{Lancaster A at SemEval-2017 Task 5: Evaluation metrics matter: predicting sentiment from financial news headlines}

\author{Andrew Moore and Paul Rayson \\
  School of Computing and Communications, Lancaster University, Lancaster, UK \\
  {\tt initial.surname@lancaster.ac.uk} 
  }

\date{}

\begin{document}
\maketitle
\begin{abstract}
This paper describes our participation in Task 5 track 2 of SemEval 2017 to predict the sentiment of financial news headlines for a specific company on a continuous scale between -1 and 1. We tackled the problem using a number of approaches, utilising a Support Vector Regression (SVR) and a Bidirectional Long Short-Term Memory (BLSTM). We found an improvement of 4-6\% using the LSTM model over the SVR and came fourth in the track. We report a number of different evaluations using a finance specific word embedding model and reflect on the effects of using different evaluation metrics.
\end{abstract}

\section{Introduction}
The objective of Task 5 Track 2 of SemEval \shortcite{semeval20175} was to predict the sentiment of news headlines with respect to companies mentioned within the headlines. This task can be seen as a finance-specific aspect-based sentiment task \cite{nasukawa2003sentiment}. The main motivations of this task is to find specific features and learning algorithms that will perform better for this domain as aspect based sentiment analysis tasks have been conducted before at SemEval \cite{pontiki2014semeval}. 

Domain specific terminology is expected to play a key part in this task, as reporters, investors and analysts in the financial domain will use a specific set of terminology when discussing financial performance. Potentially, this may also vary across different financial domains and industry sectors.
Therefore, we took an exploratory approach and investigated how various features and learning algorithms perform differently, specifically SVR and BLSTMs. We found that BLSTMs outperform an SVR without having any knowledge of the company that the sentiment is with respect to.
For replicability purposes, with this paper we are releasing our source code\footnote{\url{https://github.com/apmoore1/semeval}} and the finance specific BLSTM word embedding model\footnote{\url{https://github.com/apmoore1/semeval/tree/master/models/word2vec_models}}.



\section{Related Work}
\label{sec:related work}
There is a growing amount of research being carried out related to sentiment analysis within the financial domain. This work ranges from domain-specific lexicons \cite{loughran2011liability} and lexicon creation \cite{moore2016domain} to stock market prediction models \cite{peng2016leverage,kazemian1evaluating}. \citet{peng2016leverage} used a multi layer neural network to predict the stock market and found that incorporating textual features from financial news can improve the accuracy of prediction. \citet{kazemian1evaluating} showed the importance of tuning sentiment analysis to the task of stock market prediction. 
However, much of the previous work was based on numerical financial stock market data rather than on aspect level financial textual data. 

In aspect based sentiment analysis, there have been many different techniques used to predict the polarity of an aspect as shown in SemEval-2016 task 5 \cite{pontiki2014semeval}. The winning system \cite{brun2016xrce} used many different linguistic features and an ensemble model, and the runner up \cite{kumar2016iit} used uni-grams, bi-grams and sentiment lexicons as features for a Support Vector Machine (SVM). Deep learning methods have also been applied to aspect polarity prediction. \citet{ruder2016hierarchical} created a hierarchical BLSTM with a sentence level BLSTM inputting into a review level BLSTM thus allowing them to take into account inter- and intra-sentence context. They used only word embeddings making their system less dependent on extensive feature engineering or manual feature creation. This system outperformed all others on certain languages on the SemEval-2016 task 5 dataset \cite{pontiki2014semeval} and on other languages performed close to the best systems. \citet{wangattention} also created an LSTM based model using word embeddings but instead of a hierarchical model it was a one layered LSTM with attention which puts more emphasis on learning the sentiment of words specific to a given aspect.

\section{Data}
\label{sec:data}
The training data published by the organisers for this track was a set of headline sentences from financial news articles where each sentence was tagged with
the company name (which we treat as the aspect) and the polarity of the sentence with respect to the company. There is the possibility that the same sentence occurs more than once if there is more than one company mentioned. The polarity was a real value between -1 (negative sentiment) and 1 (positive sentiment).

We additionally trained a word2vec \cite{mikolov2013efficient} word embedding model\footnote{For reproducibility, the model can be downloaded, however the articles cannot be due to copyright and licence restrictions.} on a set of 189,206 financial articles containing 161,877,425 tokens, that were manually downloaded from Factiva\footnote{\url{https://global.factiva.com/factivalogin/login.asp?productname=global}}. The articles stem from a range of sources including the Financial Times and relate to companies from the United States only. We trained the model on domain specific data as it has been shown many times that the financial domain can contain very different language.


\section{System description}
\label{sec:system description}
Even though we have outlined this task as an aspect based sentiment task, this is instantiated in only one of the features in the SVR. The following two subsections describe the two approaches, first SVR and then BLSTM. Key implementation details are exposed here in the paper, but we have released the source code and word embedding models  to aid replicability and further experimentation.

\subsection{SVR}
The system was created using ScitKit learn \cite{pedregosa2011scikit} linear Support Vector Regression model \cite{drucker1997support}. We experimented with the following different features and parameter settings:

\subsubsection{Tokenisation}
For comparison purposes, we tested whether or not a simple whitespace tokeniser can perform just as well as a full tokeniser, and in this case we used Unitok\footnote{\url{http://corpus.tools/wiki/Unitok}}.

\subsubsection{N-grams}
We compared word-level uni-grams and bi-grams separately and in combination.

\subsubsection{SVR parameters}
We tested different penalty parameters C and different epsilon parameters of the SVR.

\subsubsection{Word Replacements}
We tested replacements to see if generalising words by inserting special tokens would help to reduce the sparsity problem. We placed the word replacements into three separate groups:
\begin{enumerate}
\item Company - When a company was mentioned in the input headline from the list of companies in the training data marked up as aspects, it was replaced by a company special token.
\item Positive - When a positive word was mentioned in the input headline from a list of positive words (which was created using the \textit{N} most similar words based on cosine distance) to `excellent' using the pre-trained word2vec model.
\item Negative - The same as the positive group however the word used was `poor' instead of `excellent'.
\end{enumerate}

In the positive and negative groups, we chose the words `excellent' and `poor' following \citet{turney2002thumbs} to group the terms together under non-domain specific sentiment words.
\subsubsection{Target aspect}
In order to incorporated the company as an aspect, we employed a boolean vector to represent the sentiment of the sentence. This was done in order to see if the system could better differentiate the sentiment when the sentence was the same but the company was different.

\subsection{BLSTM}
We created two different Bidirectional \cite{graves2005framewise} Long Short-Term Memory \cite{hochreiter1997long} using the Python Keras library \cite{chollet2015keras} with tensor flow backend \cite{abadi2016tensorflow}. We choose an LSTM model as it solves the vanishing gradients problem of Recurrent Neural Networks. We used a bidirectional model as it allows us to capture information that came before and after instead of just before, thereby allowing us to capture more relevant context within the model. Practically, a BLSTM is two LSTMs one going forward through the tokens the other in reverse order and in our models concatenating the resulting output vectors together at each time step. 

The BLSTM models take as input a headline sentence of size \textit{L} tokens\footnote{Tokenised by Unitok} where \textit{L} is the length of the longest sentence in the training texts. Each word is converted into a 300 dimension vector using the word2vec model trained over the financial text\footnote{See the following link for detailed implementation details \url{https://github.com/apmoore1/semeval\#finance-word2vec-model}}. Any text that is not recognised by the word2vec model is represented as a vector of zeros; this is also used to pad out the sentence if it is shorter than \textit{L}.

\begin{figure}[t!]
\centering
    \includegraphics[width=0.4\textwidth]{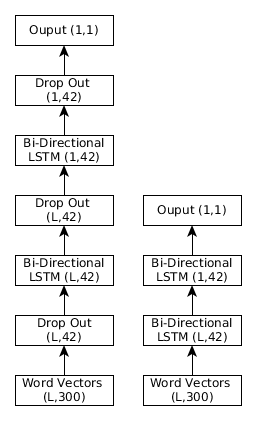}
\caption{Left hand side is the ELSTM model architecture and the right hand side shows the SLSTM. The numbers in the parenthesis represent the size of the output dimension where \textit{L} is the length of the longest sentence.}
\label{fig:model_archs}
\end{figure}

Both BLSTM models have the following similar properties:
\begin{enumerate}
\item Gradient clipping value of 5 - This was to help with the exploding gradients problem. 
\item Minimised the Mean Square Error (MSE) loss using RMSprop with a mini batch size of 32.
\item The output activation function is linear.
\end{enumerate}
The main difference between the two models is the use of drop out and when they stop training over the data (epoch). Both models architectures can be seen in figure \ref{fig:model_archs}.

\subsubsection{Standard LSTM (SLSTM)}
The BLSTMs do contain drop out in both the input and between the connections of 0.2 each. Finally the epoch is fixed at 25.

\subsubsection{Early LSTM (ELSTM)}
As can be seen from figure \ref{fig:model_archs}, the drop out of 0.5 only happens between the layers and not the connections as in the SLSTM. Also the epoch is not fixed, it uses early stopping with a patience of 10. We expect that this model can generalise better than the standard one due to the higher drop out and that the epoch is based on early stopping which relies on a validation set to know when to stop training.

\section{Results}
\label{sec:results}
We first present our findings on the best performing parameters and features for the SVRs. These were determined by cross validation (CV) scores on the provided training data set using cosine similarity as the evaluation metric.\footnote{All the cross validation results can be found here \url{https://github.com/apmoore1/semeval/tree/master/results}} We found that using uni-grams and bi-grams performs best and using only bi-grams to be the worst. Using the Unitok tokeniser always performed better than simple whitespace tokenisation. The binary presence of tokens over frequency did not alter performance. The C parameter was tested for three values; 0.01, 0.1 and 1. We found very little difference between 0.1 and 1, but 0.01 produced much poorer results. The eplison parameter was tested for 0.001, 0.01 and 0.1 the performance did not differ much but the lower the higher the performance but the more likely to overfit. 
Using word replacements was effective for all three types (company, positive and negative) but using a value \textit{N}=10 performed best for both positive and negative words. 
Using target aspects also improved results. 
Therefore, the best SVR model comprised of: Unitok tokenisation, uni- and bi- grams, word representation, C=0.1, eplison=0.01, company, positive, and negative word replacements and target aspects.

\begin{equation}
\label{eq:first_eval}
\begin{gathered}
\frac{\sum_{n=1}^{N} \text{Cosine similarity}(\hat y_n, y_n)}{N}
\end{gathered}
\end{equation}

The main evaluation over the test data is based on the best performing SVR and the two BLSTM models once trained on all of the training data. The result table \ref{tab:results} shows three columns based on the three evaluation metrics that the organisers have used. Metric 1 is the original metric, weighted cosine similarity (the metric used to evaluate the final version of the results, where we were ranked 5th; metric provided on the task website\footnote{\url{http://alt.qcri.org/semeval2017/task5/index.php?id=evaluation}}). This was then changed after the evaluation deadline to equation \ref{eq:first_eval}\footnote{Where \textit{N} is the number of unique sentences, $\hat{y}_n$ is the predicted and $y_n$ are the true sentiment value(s) of all sentiments in sentence $n$.} (which we term metric 2; this is what the first version of the results were actually based on, where we were ranked 4th), which then changed by the organisers to their equation as presented in \citet{semeval20175} (which we term metric 3 and what the second version of the results were based on, where we were ranked 5th).

\begin{table}[h]
\centering
\begin{tabular}{|l|l|l|l|l}
\hline
 Model&  Metric 1&  Metric 2& Metric 3\\
 \hline
 SVR&  62.14&  54.59&  62.34 \\
 SLSTM&  72.89&  61.55&  68.64 \\
 ELSTM&  73.20&  61.98&  69.24 \\
 \hline
\end{tabular}
\caption{Results}
\label{tab:results}
\end{table}
As you can see from the results table \ref{tab:results}, the difference between the metrics is quite substantial. This is due to the system's optimisation being based on metric 1 rather than 2. Metric 2 is a classification metric for sentences with one aspect as it penalises values that are of opposite sign (giving -1 score) and rewards values with the same sign (giving +1 score). Our systems are not optimised for this because it would predict scores of -0.01 and true value of 0.01 as very close (within vector of other results) with low error whereas metric 2 would give this the highest error rating of -1 as they are not the same sign. Metric 3 is more similar to metric 1 as shown by the results, however the crucial difference is that again if you get opposite signs it will penalise more.   

We analysed the top 50 errors based on Mean Absolute Error (MAE) in the test dataset specifically to examine the number of sentences containing more than one aspect. Our investigation shows that no one system is better at predicting the sentiment of sentences that have more than one aspect (i.e. company) within them. Within those top 50 errors we found that the BLSTM systems do not know which parts of the sentence are associated to the company the sentiment is with respect to. Also they do not know the strength/existence of certain sentiment words.

\section{Conclusion and Future Work}
\label{sec:conclusion}
In this short paper, we have described our implemented solutions to SemEval Task 5 track 2, utilising both SVR and BLSTM approaches. Our results show an improvement of around 5\% when using LSTM models relative to SVR. 
We have shown that this task can be partially represented as an aspect based sentiment task on a domain specific problem. In general, our approaches acted as sentence level classifiers as they take no target company into consideration. As our results show, the choice of evaluation metric makes a great deal of difference to system training and testing. Future work will be to implement aspect specific information into an LSTM model as it has been shown to be useful in other work \cite{wangattention}.

\section*{Acknowledgements}
We are grateful to Nikolaos Tsileponis (University of Manchester) and Mahmoud El-Haj (Lancaster University) for access to headlines in the corpus of financial news articles collected from Factiva. This research was supported at Lancaster University by an EPSRC PhD studentship.

\bibliographystyle{acl_natbib}
\bibliography{ref.bib}

\begin{thebibliography}{}
\expandafter\ifx\csname natexlab\endcsname\relax\def\natexlab#1{#1}\fi

\bibitem[{Abadi et~al.(2016)Abadi, Agarwal, Barham, Brevdo, Chen, Citro,
  Corrado, Davis, Dean, Devin et~al.}]{abadi2016tensorflow}
Mart{\'\i}n Abadi, Ashish Agarwal, Paul Barham, Eugene Brevdo, Zhifeng Chen,
  Craig Citro, Greg~S Corrado, Andy Davis, Jeffrey Dean, Matthieu Devin, et~al.
  2016.
\newblock Tensorflow: Large-scale machine learning on heterogeneous distributed
  systems.
\newblock {\em arXiv preprint arXiv:1603.04467\/} .

\bibitem[{Brun et~al.(2016)Brun, Perez, and Roux}]{brun2016xrce}
Caroline Brun, Julien Perez, and Claude Roux. 2016.
\newblock Xrce at semeval-2016 task 5: Feedbacked ensemble modelling on
  syntactico-semantic knowledge for aspect based sentiment analysis.
\newblock {\em Proceedings of SemEval\/} pages 277--281.

\bibitem[{Chollet(2015)}]{chollet2015keras}
Fran\c{c}ois Chollet. 2015.
\newblock Keras.
\newblock \url{https://github.com/fchollet/keras}.

\bibitem[{Cortis et~al.(2017)Cortis, Freitas, Daudert, Huerlimann, Zarrouk, and
  Davis}]{semeval20175}
Keith Cortis, Andre Freitas, Tobias Daudert, Manuela Huerlimann, Manel Zarrouk,
  and Brian Davis. 2017.
\newblock Semeval-2017 task 5: Fine-grained sentiment analysis on financial
  microblogs and news.
\newblock {\em Proceedings of SemEval\/} .

\bibitem[{Drucker et~al.(1997)Drucker, Burges, Kaufman, Smola, Vapnik
  et~al.}]{drucker1997support}
Harris Drucker, Christopher~JC Burges, Linda Kaufman, Alex Smola, Vladimir
  Vapnik, et~al. 1997.
\newblock Support vector regression machines.
\newblock {\em Advances in neural information processing systems\/} 9:155--161.

\bibitem[{Graves and Schmidhuber(2005)}]{graves2005framewise}
Alex Graves and J{\"u}rgen Schmidhuber. 2005.
\newblock Framewise phoneme classification with bidirectional lstm and other
  neural network architectures.
\newblock {\em Neural Networks\/} 18(5):602--610.

\bibitem[{Hochreiter and Schmidhuber(1997)}]{hochreiter1997long}
Sepp Hochreiter and J{\"u}rgen Schmidhuber. 1997.
\newblock Long short-term memory.
\newblock {\em Neural computation\/} 9(8):1735--1780.

\bibitem[{Kazemian et~al.(2016)Kazemian, Zhao, and Penn}]{kazemian1evaluating}
Siavash Kazemian, Shunan Zhao, and Gerald Penn. 2016.
\newblock \href{https://doi.org/10.18653/v1/P16-1197}{Evaluating sentiment
  analysis in the context of securities trading}.
\newblock In {\em Proceedings of the 54th Annual Meeting of the Association for
  Computational Linguistics\/}. Association for Computational Linguistics,
  pages 2094--2103.
\newblock
  \href{https://doi.org/10.18653/v1/P16-1197}{https://doi.org/10.18653/v1/P16-1197}.

\bibitem[{Kumar et~al.(2016)Kumar, Kohail, Kumar, Ekbal, and
  Biemann}]{kumar2016iit}
Ayush Kumar, Sarah Kohail, Amit Kumar, Asif Ekbal, and Chris Biemann. 2016.
\newblock Iit-tuda at semeval-2016 task 5: Beyond sentiment lexicon: Combining
  domain dependency and distributional semantics features for aspect based
  sentiment analysis.
\newblock {\em Proceedings of SemEval\/} pages 1129--1135.

\bibitem[{Loughran and McDonald(2011)}]{loughran2011liability}
Tim Loughran and Bill McDonald. 2011.
\newblock When is a liability not a liability? textual analysis, dictionaries,
  and 10-ks.
\newblock {\em The Journal of Finance\/} 66(1):35--65.

\bibitem[{Mikolov et~al.(2013)Mikolov, Chen, Corrado, and
  Dean}]{mikolov2013efficient}
Tomas Mikolov, Kai Chen, Greg Corrado, and Jeffrey Dean. 2013.
\newblock Efficient estimation of word representations in vector space.
\newblock {\em arXiv preprint arXiv:1301.3781\/} .

\bibitem[{Moore et~al.(2016)Moore, Rayson, and Young}]{moore2016domain}
Andrew Moore, Paul Rayson, and Steven Young. 2016.
\newblock Domain adaptation using stock market prices to refine sentiment
  dictionaries.
\newblock In {\em Proceedings of the 10th edition of Language Resources and
  Evaluation Conference (LREC2016)\/}. European Language Resources Association
  (ELRA).

\bibitem[{Nasukawa and Yi(2003)}]{nasukawa2003sentiment}
Tetsuya Nasukawa and Jeonghee Yi. 2003.
\newblock Sentiment analysis: Capturing favorability using natural language
  processing.
\newblock In {\em Proceedings of the 2nd international conference on Knowledge
  capture\/}. ACM, pages 70--77.

\bibitem[{Pedregosa et~al.(2011)Pedregosa, Varoquaux, Gramfort, Michel,
  Thirion, Grisel, Blondel, Prettenhofer, Weiss, Dubourg
  et~al.}]{pedregosa2011scikit}
Fabian Pedregosa, Ga{\"e}l Varoquaux, Alexandre Gramfort, Vincent Michel,
  Bertrand Thirion, Olivier Grisel, Mathieu Blondel, Peter Prettenhofer, Ron
  Weiss, Vincent Dubourg, et~al. 2011.
\newblock Scikit-learn: Machine learning in python.
\newblock {\em Journal of Machine Learning Research\/} 12(Oct):2825--2830.

\bibitem[{Peng and Jiang(2016)}]{peng2016leverage}
Yangtuo Peng and Hui Jiang. 2016.
\newblock \href{https://doi.org/10.18653/v1/N16-1041}{Leverage financial news
  to predict stock price movements using word embeddings and deep neural
  networks}.
\newblock In {\em Proceedings of the 2016 Conference of the North American
  Chapter of the Association for Computational Linguistics: Human Language
  Technologies\/}. Association for Computational Linguistics, pages 374--379.
\newblock
  \href{https://doi.org/10.18653/v1/N16-1041}{https://doi.org/10.18653/v1/N16-1041}.

\bibitem[{Pontiki et~al.(2014)Pontiki, Galanis, Pavlopoulos, Papageorgiou,
  Androutsopoulos, and Manandhar}]{pontiki2014semeval}
Maria Pontiki, Dimitris Galanis, John Pavlopoulos, Harris Papageorgiou, Ion
  Androutsopoulos, and Suresh Manandhar. 2014.
\newblock Semeval-2014 task 4: Aspect based sentiment analysis.
\newblock {\em Proceedings of SemEval\/} pages 27--35.

\bibitem[{Ruder et~al.(2016)Ruder, Ghaffari, and
  Breslin}]{ruder2016hierarchical}
Sebastian Ruder, Parsa Ghaffari, and G.~John Breslin. 2016.
\newblock \href{http://aclweb.org/anthology/D16-1103}{A hierarchical model of
  reviews for aspect-based sentiment analysis}.
\newblock In {\em Proceedings of the 2016 Conference on Empirical Methods in
  Natural Language Processing\/}. Association for Computational Linguistics,
  pages 999--1005.
\newblock
  \href{http://aclweb.org/anthology/D16-1103}{http://aclweb.org/anthology/D16-1103}.

\bibitem[{Turney(2002)}]{turney2002thumbs}
Peter~D Turney. 2002.
\newblock Thumbs up or thumbs down?: semantic orientation applied to
  unsupervised classification of reviews.
\newblock In {\em Proceedings of the 40th annual meeting on association for
  computational linguistics\/}. Association for Computational Linguistics,
  pages 417--424.

\bibitem[{Wang et~al.(2016)Wang, Huang, zhu, and Zhao}]{wangattention}
Yequan Wang, Minlie Huang, xiaoyan zhu, and Li~Zhao. 2016.
\newblock \href{http://aclweb.org/anthology/D16-1058}{Attention-based lstm for
  aspect-level sentiment classification}.
\newblock In {\em Proceedings of the 2016 Conference on Empirical Methods in
  Natural Language Processing\/}. Association for Computational Linguistics,
  pages 606--615.
\newblock
  \href{http://aclweb.org/anthology/D16-1058}{http://aclweb.org/anthology/D16-1058}.

\end{thebibliography}

\appendix

\end{document}